\documentclass{article}

\PassOptionsToPackage{numbers, compress}{natbib}

\usepackage[preprint]{neurips_2024}

\usepackage{microtype}
\usepackage{subfigure}
\usepackage{booktabs} % for professional tables
\usepackage{listings}
\usepackage{multirow}
\usepackage{tabularx}
\usepackage{adjustbox}
\usepackage{graphicx}
\usepackage{amsmath}
\usepackage{amssymb}
\usepackage{mathtools}
\usepackage{amsthm}
\usepackage{algorithm}
\usepackage{algorithmic}
\usepackage[utf8]{inputenc} % allow utf-8 input
\usepackage[T1]{fontenc}    % use 8-bit T1 fonts
\usepackage{hyperref}       % hyperlinks
\usepackage{url}            % simple URL typesetting
\usepackage{booktabs}       % professional-quality tables
\usepackage{amsfonts}       % blackboard math symbols
\usepackage{nicefrac}       % compact symbols for 1/2, etc.
\usepackage{microtype}      % microtypography
\usepackage{xcolor}         % colors

\usepackage{titlesec}
\titlespacing*{\section}{0pt}{0.1\baselineskip}{0.1\baselineskip}
\titlespacing*{\subsection}{0pt}{0.1\baselineskip}{0.1\baselineskip}
\titlespacing*{\subsubsection}{0pt}{0.1\baselineskip}{0.1\baselineskip}

\title{Diffusion Model for Relational Inference}

\author{%
Shuhan Zheng$^{1,2}$\thanks{Now works in Hitachi, Ltd.} \quad Ziqinag Li$^{1}$ \quad Kantaro Fujiwara$^1$ \quad Gouhei Tanaka$^{1,2}$ \\
$^1$International Research Center for Neurointelligence (IRCN), The University of Tokyo\\ 
$^2$Graduate School of Engineering, Nagoya Institute of Technology\\
}

\begin{document}

\maketitle

\begin{abstract}
Dynamical behaviors of complex interacting systems, including brain activities, financial price movements, and physical collective phenomena, are associated with underlying interactions between the system's components. The issue of uncovering interaction relations in such systems using observable dynamics is called relational inference. In this study, we propose a \textbf{Diff}usion model for \textbf{R}elational \textbf{I}nference (DiffRI), inspired by a self-supervised method for probabilistic time series imputation. DiffRI learns to infer the probability of the presence of connections between components through conditional diffusion modeling.
%Our model chooses one of the observed multiple time series as a target, introduces randomly masked sections in the target time series, and then aims to impute missing samples in the masked sections using all the other time series based on a diffusion model while estimating the probability of the presence of interactions between the target and the other ones.By repeating the above process while changing the target time series, the whole relation between components can be inferred. To our knowledge, DiffRI is the first approach that employs time series diffusion modeling for relational inference. 
Experiments on both simulated and quasi-real datasets show that DiffRI is highly competent compared with several well-known methods in discovering ground truth interactions in an unsupervised manner. %Our code is available at \url{https://anonymous.4open.science/r/diffri-215E/}.
\end{abstract}

\section{Introduction}
\label{sec:intro}
\textbf{Background}
Machine learning-based approaches have demonstrated considerable success in computational tasks such as time series forecasting, classification, and imputation, as evidenced by a growing body of research ~\citep{fawaz2019deep, lim2021time,li2022multi,jin2023survey, kim2023probabilistic}. Despite these rapid-pace advancements, the crucial task of inferring relations between components in interacting systems from observable temporal behavior of their states is still challenging. Unearthing the underlying mechanisms of interacting systems is frequently of greater significance than simply predicting system behavior under known interaction information. In fact, inferring relations between multiple or multivariate time series offers a useful step for comprehending the whole system dynamics resulting from interacting dynamical components. For instance, neuroscientists strive to discern relations between neural signals to elucidate functional connectivity amongst various ROIs (Region of Interests) of the brain~\cite{liu2018detecting,li2022functional, zhang2023oxytocin}. Similarly, financial analysts employ causal inference methodologies to uncover cause-and-effect connections within financial time series data~\cite{antonakis2010making, moraffah2021causal}. In this way, relational inference from observed dynamics is a pervasive and fundamental challenge that spans many disciplines.

In the machine learning community, there has been a concerted effort to devise advanced techniques for relational inference from observable time series data. A pioneering contribution in this field is the Neural Relational Inference (NRI) model, introduced by Kipf et al., which integrates a Variational AutoEncoder (VAE) with Graph Neural Network (GNN) operations to construct neural representations~\cite{kipf2018neural}. NRI is designed to infer interactions of entities (or components) in an encoder part while simultaneously learning their dynamics for forecasting future states in a decoder part. Building upon this foundational framework, variants of the NRI model have emerged, including Dynamic NRI~\cite{graber2020dynamic}, NRI-MPM~\cite{chen2021neural}, and ACD~\cite{lowe2022amortized}, each aiming to refine and improve the capabilities of the original method. More recent developments include a relational inference model proposed by Wang et al., which employs reservoir networks for computationally efficient relational inference \cite{wang2023effective}. Nevertheless, most relational inference models developed so far are based on a VAE and trained in the framework of time series prediction.

Recently, a family of diffusion models (or score-based generative models)~\cite{ho2020denoising,song2020denoising,song2020score} following the original one \cite{sohl2015deep} have shown excellent performance in various tasks, ranging from multimodal information processing~\cite{rombach2022high, saharia2022photorealistic} to tabular data imputation~\cite{zheng2022diffusion, kotelnikov2023tabddpm}. Successful applications in practice suggest that the diffusion model can accurately estimate high-dimensional data distribution through score matching~\cite{yang2023diffusion}. On the theoretical side, Oko et al. showed that under certain assumptions, compared with the true distribution, the generated data distribution achieves the nearly minimax optimal estimation rates in the total variation distance and in the Wasserstein distance of order one~\cite{oko2023diffusion}. 

\textbf{Motivations} The powerful computational ability of the diffusion models inspired us to evaluate its potential in relational inference tasks. Previously, \textit{prediction-based VAE framework} (as developed in the seminal NRI work~\cite{kipf2018neural}) stands for the milestone neural network-based approach for relational inference in time series data. Our study explores the possibility of using the \textit{imputation-based diffusion} approach for relational inference from time series data for the first time. First, diffusion-based models show superiority in various time series data processing~\cite{yang2023diffusion}, providing evidence for strong ability in modeling time series. For example, experimental results in time series imputation tasks show that diffusion-based imputation models perform better than VAE-based models~\cite{tashiro2021csdi,alcaraz2023diffusionbased}. Second, the imputation-based training method is more suitable than the prediction-based training method for datasets containing missing data. Models trained through prediction assume past time steps to be observed~\cite{kipf2018neural, chen2021neural,wang2023active}, while the imputation model does not need this assumption. Therefore, for datasets with missing data, it is better to train the model through imputation rather than prediction, as illustrated in Sec.~\ref{sec:results-ablation studies} %These results motivated us to develop a diffusion-based model for neural relational inference using multiple time series data observed from interacting dynamical systems. \textbf{How about dataset size requirement advantage? I think diff will require less data than VAE approach.}

The motivation led us to propose a novel time series generative model: a \textbf{Diff}usion model for \textbf{R}elational \textbf{I}nference (DiffRI). Specifically, we designed a diffusion-based model that combines self-supervised time series imputation with relational learning. Our main contributions are as follows:
\begin{itemize}
  \item To our knowledge, DiffRI is the first model that attempts neural networks-based relational inference on time series through diffusion generative modeling. We demonstrate the potential of diffusion models in relational inference tasks. 
  \item The proposed method can infer relations within a set of multiple time series data with high accuracy across different interaction mechanisms. Moreover, DiffRI does not need additional data augmentation to achieve good results and is robust to missing data.
  \item The proposed method can correctly identify relations in a quasi-real neural signal dataset with realistic settings without any regularization techniques.
\end{itemize}
The paper is organized as follows. In Sec~\ref{sec:relatedworks}, we mention related works on relational inference. In Sec~\ref{sec:problem_form}, we formulate the relational inference problem. In Sec~\ref{sec:method}, we explain our methodology utilizing diffusion models. In Sec~\ref{sec:experiments}, we show experimental results on the performance of the proposed model while comparing it with several well-known models.

\section{Related works}
\label{sec:relatedworks}
A variety of methods have been developed to address the relational inference problem with time series data. We can categorize these methods into three types: model-free descriptive statistics-based (e.g., correlation), model-free information theory-based (e.g., mutual information), and model-based methods (e.g., autoregressive models)~\cite{de2018connectivity,brugere2018network}. Here, we focus on model-based methods implemented with neural networks in machine learning frameworks. %Other model-based models often assume deterministic or stochastic dynamical systems models that can reproduce observational data by fitting model parameters~\cite{de2018connectivity}. 

Most model-based methods~\cite{kipf2018neural,chen2021neural,banijamali2021neural} are grounded in the NRI model proposed by \cite{kipf2018neural}, which integrates a VAE with GNN operations to construct neural representations and learns to predict future dynamics. For example, Chen et al. introduced efficient message-passing mechanisms into the GNNs with structural prior knowledge~\cite{chen2021neural}. Later, Banijamali further improved the NRI model by incorporating individualized information with newly introduced private nodes in the graph~\cite{banijamali2021neural}. Recently, using GNN to capture relations in time series data has become popular~\cite{zhang2021graph,liu2022multivariate}. However, there is no experimental evidence showing that the learned GNN can be reliably and easily used for relational inference tasks.

Extended frameworks derived from the NRI model have been presented. For example, Zhang et al. introduced a general framework, Gumbel Graph Network (GGN), consisting of a network generator and a dynamics learner for performing both network interaction inference and time series prediction, which are jointly trained with backpropagation~\cite{zhang2019general}. Wang et al. proposed a novel integration of a Reservoir Computing (RC) model with a VAE for computationally efficient structural inference~\cite{wang2023effective}. Additionally, they explored an active learning strategy combined with Partial Information Decomposition (PID)~\cite{wang2023active}. 

Other inspiring related works include papers by Cini et al., and Cheng et al.~\cite{cheng2022cuts,cini2023sparse}. Cini et al. introduced a graph learning module to optimize prediction performance. Cheng et al.'s research introduced an imputation stage, which boosts the subsequent causal discovery stage. 

We notice that most related works adopt a prediction-based model or the VAE framework. In this work, we present a novel solution to relational inference from time series data. Our work leverages a diffusion model for time series imputation in combination with an interaction estimation process.

\section{Problem formulation}
\label{sec:problem_form}
We give a mathematical formulation of the relational inference problem using observational time series data from a dynamical system of interacting components. Let $\mathbf{X} \in \mathbb{R}^{K\times L}$ be multivariate time series data representing the time evolution of the states of the system's components, where $K$ denotes the number of components and $L$ denotes the length of time series. Note that for multidimensional features, we sequentially concatenate different features in temporal dimension $L$.

A static connectivity between the $K$ components can be represented as a fixed graph $G = \{\mathcal{V}, \mathcal{E}\}$ with vertices (or nodes) $v\in \mathcal{V}$ and edges $e \in \mathcal{E}$, which correspond to the system's components and their connectivity, respectively. %We assume that the temporal evolution of the states on the vertices are observable as time series data. That is,
A directed edge $e=(v_i,v_j) \in \mathcal{E}$ from vertex $v_i$ to vertex $v_j$ means that the time evolution of the state at $v_i$ influences that at $v_j$.

The transductive problem focused in this work is to infer a set of latent edges $\mathcal{E}$ in the underlying graph $G$ when given a set of multivariate time series data. 

\section{Method}
\label{sec:method}
This section introduces DiffRI, a conditional generative model used to infer relations between interacting components from time series data. DiffRI is constructed on the basis of a conditional time series diffusion model~\cite{tashiro2021csdi} that separates observed multivariate time series into an imputation target time series and the other conditional time series. We let the model select informative time series from the conditional part and perturb non-informative parts. Different from prediction-based models, DiffRI is trained to impute masked parts in the target time series. The concept and overview of the DiffRI framework are presented in Sec.~\ref{sec:framework}. More detailed explanations are provided in Secs.~\ref{sec:edgepredmod}-\ref{sec:inference}.

%We note that each row of the multivariate time series data $\mathbf{X}$ corresponds to a vertex (or node) of a latent graph. We also note that, unlike many previous models with the GNN formalism for relational inference, DiffRI is not permutation invariant. This means that the ordering of the rows in the data matrix must be preserved in DiffRI. 

\subsection{DiffRI framework}
\label{sec:framework}   

\begin{figure*}[htbp]
  \centering
  \includegraphics[width=\textwidth]{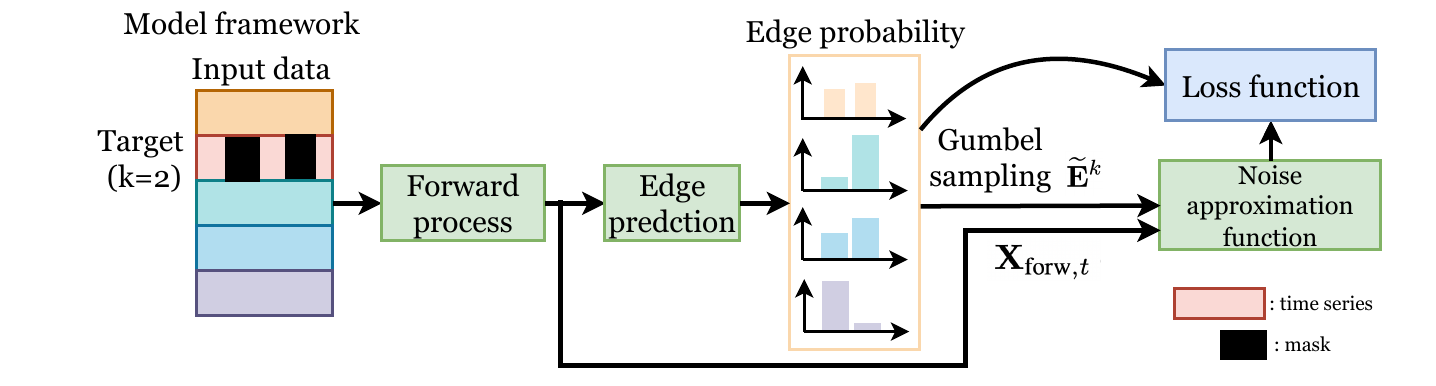}
  \caption{An illustration of DiffRI framework when $K$ observational time series with length $L$ are given. The figure shows an example when $K=5$ and $k=2$ (as in the other figures). Imputation targets (i.e. the masked sections in the $k$-th time series at target node $k$) become noisy in the forward diffusion process. Matrix $\mathbf{X}_{\text{forw,t}} \in \mathbb{R}^{K\times L}$ denotes the set of time series obtained at the diffusion step $t$ in the forward process. The output of the edge prediction module, $\widetilde{\mathbf{E}}^k$, indicates sampled edges associated with the target node, which is used in the noise approximation process.}
  \label{fig:framework}
  \vspace{-0.15in}

\end{figure*}

The DiffRI model consists of modules for a forward diffusion process, a reverse denoising (or noise approximation) process, and an edge prediction process as illustrated in Figure~\ref{fig:framework}. 

First, inspired by the Conditional Score-based Diffusion model for Imputation (CSDI)~\cite{tashiro2021csdi}, we divide the time series data $\mathbf{X} \in \mathbb{R}^{K\times L}$ into imputation targets $\mathbf{x}_0^k$ and conditional observations $\mathbf{X}^{\text{co}}$, where $k$ denotes the index of the target node. This setting is employed to consider possible interactions between the target node and the other ones. The imputation targets are determined by random masking as detailed in Appx.~\ref{appen:masking_strategy}. 

The diffusion process is conditional. Therefore, only $\mathbf{x}_0^k$ go through the forward diffusion process, while $\mathbf{X}^{\text{co}}$ are kept unchanged to serve as conditional data in the reverse denoising process. To describe these processes, we define $\mathbf{x}^{k}_t$ for $t=1,\ldots,T$ as a sequence of latent variables in the same sample space as imputation target $\mathbf{x}^k_0$. In multidimensional cases, we adapt $\mathbf{x}^k_0$ and $\mathbf{X}^{\text{co}}$ by concatenating features from different dimensions. We formulate the forward process as the following Markov chain:
\begin{align}
\begin{split}
q(\mathbf{x}^k_{1:T}|\mathbf{x}^k_0) &:= \prod_{t=1}^T q(\mathbf{x}^k_t|\mathbf{x}^k_{t-1}), \\
q(\mathbf{x}_{t}^{k}|\mathbf{x}_{t-1}^{k}) &:= \mathcal{N}(\mathbf{x}_{t}^k;\sqrt{1-\beta_{t}}\mathbf{x}^{k}_{t-1}, \beta_t \mathbf{I}),
\end{split}
\label{eq:forw}
\end{align}
and the reverse process as the following Markov chain:
\begin{align}
& p_{\theta}(\mathbf{x}^k_{0:T}|\mathbf{X}^{\text{co}}) := p(\mathbf{x}^k_T)\prod_{t=1}^T p_\theta(\mathbf{x}^k_{t-1}|\mathbf{x}^k_t,\mathbf{X}^{\text{co}},\mathbf{E}^k), \nonumber \\
&\mathbf{x}_T^k \sim \mathcal{N}(\mathbf{0}, \mathbf{I}),\\
& p_{\theta}(\mathbf{x}^{k}_{t-1}|\mathbf{x}^{k}_{t},\mathbf{X}^{\text{co}},\mathbf{E}^k) := \mathcal{N}(\mathbf{x}^{k}_{t};\mathbf{\mu}_\theta(\mathbf{x}^{k}_{t}, \mathbf{X}^{\text{co}}, \mathbf{E}^k,t), \sigma^2_t \mathbf{I}). \nonumber
\label{eq:back}
\end{align} 

The pre-defined parameters $\beta_t$ and $\sigma_t$ are determined following the settings in CSDI~\cite{tashiro2021csdi}, and $\mathbf{E}^k$ represents an edge existence probability produced by the edge prediction module (see Sec.~\ref{sec:edgepredmod} for details). We randomly sample $k$ from $\mathbb{N}_{\leq K}$ and $t$ from $\mathbb{N}_{\leq T}$ during model training, where $T$ is the total number of diffusion steps. %\sout{Conditional data $\mathbf{X}^{\text{co}} \in \mathbb{R}^{K\times L}$ represents data sample excluding $\mathbf{x}^{k}_t$.} 
We denote a matrix combining $\mathbf{x}^k_t$ and $\mathbf{X}^{\text{co}}$ at diffusion time step $t$ by $\mathbf{X}_{\text{forw},t} \in \mathbb{R}^{K\times L}$.

Following the design of Denoising Diffusion Probabilistic Model (DDPM) proposed by \cite{ho2020denoising}, $\mathbf{\mu}_\theta$ can be further parameterized as follows:
\begin{align}
\mu_\theta=\frac{1}{\alpha_t}\left(\mathbf{x}^{k}_{t-1}-\frac{\beta_t}{\sqrt{1-\alpha_t}}\epsilon^k_\theta(\mathbf{x}^{k}_{t}, \mathbf{X}^{\text{co}}, \mathbf{E}^k, t) \right),
\end{align}
where $\epsilon^k_\theta$ is a trainable noise approximation (or denoising) function, corresponding to a rescaled score model in the context of score-based generative models. All the other parameters are pre-defined. We adapt the DiffWave model architecture~\cite{kong2020diffwave} as the backbone of $\epsilon^k_\theta$. The DiffWave architecture is modified by introducing a feature interaction layer and a perturbation operator (see Sec. ~\ref{sec:feature_inter} for details).

Then, we train DiffRI by minimizing the denoising loss function $||\epsilon^k-\epsilon^k_\theta||^2$, where $\epsilon^k$ is the ground truth noise, with optional regularization losses (see Sec.~\ref{sec:reg_loss} for details). The trained model is used for relational inference (see Sec.~\ref{sec:inference} for details). 

We provide a theoretical interpretation for DiffRI in Appx.~\ref{appen:kldiv}-\ref{appen:theo_inter}. We show that the essential role of DiffRI is to seek time series that can minimize expected message length for $p_\theta$ to approximate the ground truth reverse probability distribution. Our theoretical analysis supports the validity of the proposed framework and numerical results in Sec.~\ref{sec:experiments} demonstrate its effectiveness.

\subsection{Edge prediction module}
\label{sec:edgepredmod}

We represent the edge prediction module as a function $g(\mathbf{X}_{\text{forw},t},k): \mathbb{R}^{K\times L}\times \mathbb{N}_{\leq K}\to\mathbb{R}^{(K-1)\times 2}$. The edge existence probability is determined by the multivariate time series $\mathbf{X}_{\text{forw},t} \in \mathbb{R}^{K\times L}$ at diffusion step $t$ and the target node index $k\in \mathbb{N}_{\leq K}$. 
%Matrix $\mathbf{X}_{\text{forw},t}$ contains both $\mathbf{X}^{\text{co}}$ and $\mathbf{x}^{k}_t$. 
In the following texts, we omit $t$ representing the diffusion step when the meaning is obvious. %These logits are computed with the Gumbel-softmax technique~\cite{45822} as described later.
%In DiffRI, we expect $g(\cdot,\cdot)$ to give a set of $\{\mathbf{E}^{k}, k=1,..., K\}$ that faithfully represent ground truth interactions between time series.

\begin{figure*}[htbp]
  \centering
  \includegraphics[width=\textwidth]{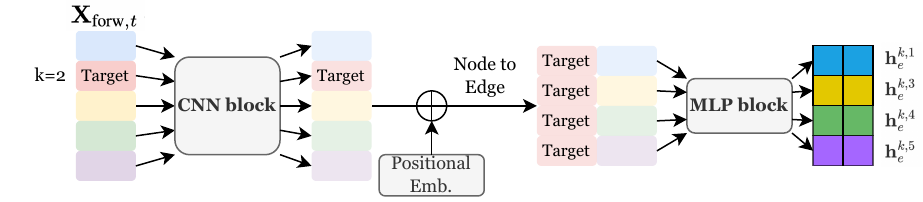}
  \caption{An example of edge prediction module. Initially, each time series goes through a CNN block. Then, positional embeddings are added to the feature maps in order to incorporate information regarding ordering. Subsequently, a node-to-edge transformation is applied to obtain edge representations that encapsulate the relations between node pairs. Finally, edge representations are fed into an MLP block. The output of the MLP block is given as logits $\mathbf{h}^{k,i}_e$ for $i\in \{\mathbb{N}_{\leq K}| i\neq k\}$, which determine the edge probability between target node $k$ and other node $i$.}
  \label{fig:edgepred}
  \vspace{-0.0in}
\end{figure*}

In this work, we implement the edge prediction module using neural networks as illustrated in Fig.~\ref{fig:edgepred}. The process is described as follows:
\begin{align}
    \mathbf{H}_{\text{emb}}&=f_{\text{CNN}}(\mathbf{X}_{\text{forw},t}) + \mathbf{S}_{pos}, \label{eq:Hemb}\\
    \mathbf{h}^{k, i}_e &= f_{\text{Linear}}(f_{\text{MLP}}([\mathbf{h}^{k}_{\text{emb}}, \mathbf{h}^{i}_{\text{emb}}])), i\in \mathbb{N}_{\leq K}, i\neq k,\label{eq:eki} 
\end{align}
where $f_{\text{CNN}}, f_{\text{MLP}}$ and $f_{\text{Linear}}$ represent a convolutional neural network (CNN), a multilayer perceptron (MLP), and a linear matrix multiplication (Linear), respectively. In Eq.~(\ref{eq:Hemb}), $\mathbf{X}_{\text{forw},t}$ is first transformed by a CNN-based block into feature maps, to which a positional embedding tensor $\mathbf{S}_{pos}$ is added. The positional embedding tensor is introduced to strengthen the model's ability to distinguish different time series~\cite{vaswani2017attention}. Then, we obtain $\mathbf{H}_{\text{emb}} =[\mathbf{h}^{1}_{\text{emb}},...,\mathbf{h}^{K}_{\text{emb}}] \in \mathbb{R}^{K\times H}$  
%which gives node embedding for each time series. 
where $\mathbf{h}^{j}_{\text{emb}}$ is an embedding vector associated with node $j$ for $1\le j\le K$ and $H$ is the number of hidden units. In Eq.~(\ref{eq:eki}), we concatenate $\mathbf{h}^{k}_{\text{emb}}$ with $\mathbf{h}^{i}_{\text{emb}} (i\in \{\mathbb{N}_{\leq K}| i\neq k\})$ to denote an edge representation between target node $k$ and another node $i$. Lastly, we apply an MLP block and a Linear layer to project the edge representation to logits $\mathbf{h}^{k,i}_e \in \mathbb{R}^{2}$. $\{\mathbf{h}^{k,i}_e\in \mathbb{R}^2; i\in N_{\le K}, i\neq k\}$ contains a pair of logits for edge existence and non-existence from node $i$ to the target node $k$ for each $i$.

Let $\mathbf{e}^{k,i}\in\mathbb{R}^{2}$ be a random variable where the two elements (each taking 0 or 1) indicate edge existence and non-existence between node $k$ and $i$. Following Eq.~(\ref{eq:eki}), to obtain edge existence probability distribution $p'_\theta(\mathbf{e}^{k,i}|\mathbf{X}_{\text{forw,t}},k)$, we apply the softmax operation to the logits as follows:
\begin{align}
p'_\theta(\mathbf{e}^{k,i}|\mathbf{X}_{\text{forw,t}},k)& =\text{softmax}(((\mathbf{h}^{k,i}_e)^2+\eta)/ \tau)\in \mathbb{R}^{2}, \nonumber \\
&i\in \mathbb{N}_{\leq K}, i\neq k,
    %p_\theta(\mathbf{E}^k|\mathbf{X}_{\text{forw,t}},k) &= [\mathbf{z}_{k,1},...,\mathbf{z}_{k,k-1},\mathbf{z}_{k,k+1},...,\mathbf{z}_{k,K}],
\end{align}
where $\tau$ is the softmax temperature parameter and $\eta \sim \text{Gumbel}(0,1)$. By using the famous Gumbel-softmax sampling trick~\cite{45822}, we obtain a discrete edge sample $\widetilde{\mathbf{e}}^{k,i} \in \{[0,1],[1,0] \}$ from the discrete probability $p_\theta'(\mathbf{e}^{k,i}|\mathbf{X}_{\text{forw,t}},k)$. If the first (resp. second) element of $\widetilde{\mathbf{e}}^{k,i}$ equals $1$, it indicates edge existence (resp. non-existence).
\subsection{Feature interaction layer with perturbation operator}
\label{sec:feature_inter}
In this section, we first briefly introduce the architecture of the noise approximation model and then focus on the feature interaction layer with the perturbation operator. 

Firstly, about the backbone of the noise approximation model, we followed the DiffWave architecture adopted in the CSDI model~\cite{tashiro2021csdi}. The adopted DiffWave model predicts noise $\epsilon^k$ in the target data $\mathbf{x}_t^k$. For more details, see Appx.~\ref{appenx:Neural network implementation of the noise approximation function}.

\begin{figure}[htbp]
  \centering
  \includegraphics[]{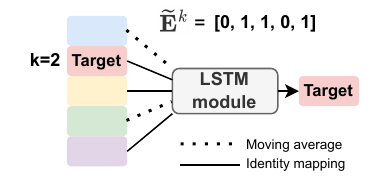}
  \caption{An example of feature interaction layer based on an LSTM module and a moving average perturbation operation. Following sampled $\widetilde{\mathbf{E}}^k$, we aggregate time series to update the target time series (red block) by the LSTM module. The time series corresponding to node $i=1,4$, which are considered to have no interactions with node $k$, will undergo the moving average operation.}
  \label{fig:feature_inter}
  \vspace{-0.1in}
\end{figure}

Secondly, in DiffRI, we introduce an LSTM-based feature interaction layer combined with a moving average perturbation operator~\cite{crabbe2021explaining}. Our designs are illustrated in Fig.~\ref{fig:feature_inter}. We formulate our approach as follows:
\begin{align}
\textbf{X}_{\text{MA}} &= \pi_{\text{MA}}((1-\widetilde{\mathbf{E}}^{k}) \odot \mathbf{X}_{\text{temporal}})\label{equ:X_ma}, \\
\textbf{x}^{k}_{\text{feat}} &=f_{\text{LSTM}}(\textbf{X}_{\text{MA}}+\widetilde{\mathbf{E}}^{k}\odot \mathbf{X}_{\text{temporal}})\label{equ:x_feat^k},
\end{align}
where $\mathbf{X}_{\text{temporal}}\in \mathbb{R}^{K\times L\times C}$ represents time series passing through the temporal transformer layer, $C$ denotes the number of channels (aka. embedding dimension),  $\pi_{\text{MA}}$ denotes moving average along temporal dimension, $\widetilde{\mathbf{E}}^{k}=[\widetilde{\mathbf{e}}^{k,1}[0], \widetilde{\mathbf{e}}^{k,2}[0], ..., \widetilde{\mathbf{e}}^{k,K}[0]]\in \{0,1\}^K$ (treated as a $1\times K$ matrix) denotes sampled edges, and $\odot$ denotes element-wise broadcast multiplication. We assign a value of $1$ to $\widetilde{\mathbf{E}}^k[k]$ to signify self-connection.

In Eq.~(\ref{equ:X_ma}), we apply the moving average as the perturbation operator to time series corresponding to node indices in $\{i~|~\widetilde{\mathbf{E}}^k[i]=0\}$ that are considered to have no interactions with target node $k$. The perturbation operator stands for a clever way to “blur” information. The perturbation operator perturbs data to reduce the impact of unconnected nodes. However, the perturbation operation still allows data to pass through. This means that if the model initially makes an incorrect relational inference, it has the flexibility to correct itself. In Eq.~(\ref{equ:x_feat^k}), we apply an LSTM module to the addition of $\mathbf{X}_\text{MA}$ and unperturbed interacting time series data $\widetilde{\mathbf{E}}^{k}\odot \mathbf{X}_{\text{temporal}}$. The LSTM block processes the information of features with dimension $K$ sequentially, which respects causal relation. We use the LSTM hidden state at each time step (along with channel) in the sequence to update the representation of the target time series $\mathbf{x}_{\text{feat}}^k \in \mathbb{R}^{L\times C}$. %Our design is illustrated in Fig.~$\ref{fig:feature_inter}$.

%In Fig.~$\ref{fig:feature_inter}$, the five color blocks depict five time series given to the feature interaction layer, and the red block corresponds to the target node $k=2$. Based on the inferred edge information $\widetilde{\mathbf{E}}^k$, we update the representation of the target parts by applying an LSTM module to the corresponding time series. %In addition, we provide an alternative MLP module for feature interaction learning.

Following the LSTM block, the updated time series is fed into subsequent parts of DiffWave architecture to give noise approximator $\epsilon^k_\theta$. For more information about our implementation of the noise approximation model, refer to Appx.~\ref{appenx:Neural network implementation of the noise approximation function}.
\subsection{Training with a regularization loss}
\label{sec:reg_loss}

In our study, we incorporate network density $\rho \in (0,1]$ as an \textit{optional} structural prior information into our objective function. Network density is defined as the proportion of actual connections (or interactions) between node pairs to all possible ones in the whole network. Together with noise approximation loss, we formulate our training objective function as follows:
\begin{equation}
\begin{aligned}
    \min_\theta\mathcal{L}(\theta) &=\min_{\theta}\mathbb{E}_{\epsilon\sim\mathcal{N}(0,1),t,k}\lVert\epsilon^k-\epsilon^k_\theta(\mathbf{x}^{k}_{t}, \mathbf{X}^{\text{co}}, \mathbf{E}^k, t)\rVert_2^2\\
    &+\lambda_1 \mathbb{E}_{k} \left\| \frac{\sum_{i,i\neq k}{\mathbf{E}^{k}[i]}}{K-1}-\rho \right\|,
    \label{equ:loss}
\end{aligned}
\end{equation}
where $\textbf{X}\in \mathbb{R}^{K\times L}$ denotes a multivariate time series sample, $k$ is the index of target node, $t$ is the diffusion step, the variable $\mathbf{E}^{k}[i]$ denotes the edge presence/absence between target node $k$ and other nodes $i$, $\lambda_1$ is the coefficient of the regularization term, and $\rho$ is the network density. Again, we emphasize that the regularization loss is \textit{optional}. In Sec.~\ref{sec:mainresults_netsim} and Sec.~\ref{sec:results-ablation studies}, we found that DiffRI can also perform well when the regularization term is omitted during training.

\subsection{Inference with DiffRI}
\label{sec:inference}
After training the model, we run DiffRI for relational inference. We set each time series as the target time series one by one and run the trained model to denoise imputation target parts within the target time series. %Within the reverse process, we collect the inferred edges and sum up the results. Finally, we use network density as a threshold to determine whether edges exist between nodes. 

\begin{figure}[htbp]
  \centering
  \includegraphics[width=8.8cm,height=3.8cm]{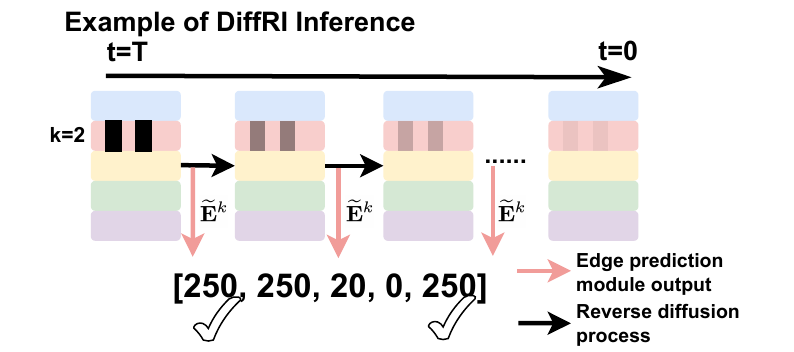}
  \caption{An example of DiffRI inference. The color blocks symbolize time series, while the black blocks denote noisy parts. We accumulate the sampled edge $\widetilde{\mathbf{E}}^k$ during each reverse diffusion step to get a score indicating the potential connection between target nodes and remaining nodes. Given prior structural knowledge, edges scoring in the top 100$\rho$ percentile are considered to interact with the target node (marked with \checkmark)}
  \label{fig:inference}
  \vspace{-0.1in}
\end{figure}

Fig.~\ref{fig:inference} provides an example case of inferring incoming edges to target node $k(=2)$. Random sections in the $k$-th time series are set as imputation targets (selected by a mask). Through the reverse process, the edge prediction module generates sampled directed edges $\widetilde{\mathbf{E}}^k$ that point towards the target node $k$. We then sum these sampled results up and utilize the network density as structural prior knowledge to ascertain whether an edge exists. In this particular illustration of the score, DiffRI determines that directed edges from both the first and fifth nodes are pointing toward the second node. This procedure is repeated for each target time series until we finally derive all the relations (i.e. connectivity) for the whole pairs of components.

\section{Experiments}
\label{sec:experiments}
\subsection{Datasets}
In numerical experiments, we adopted four time series datasets: the Kuramoto model, spring systems, vector autoregression (VAR) models, and NetSim, a quasi-real dataset describing the fMRI signals of human brain regions~\cite {smith2011network}. Here, we provide brief descriptions of the four datasets  (see Appx.~\ref{appen:datasets} for details). The performance of DiffRI was examined in relational inference tasks. \\
\textbf{Kuramoto}:
The Kuramoto model ~\cite{kuramoto1975self}, consisting of interacting phase oscillators, has been widely used in studying oscillatory behavior of complex systems, ranging from biological systems to chemical ones. \\
\textbf{Spring}:
We examined a two-dimensional (2D) physical model in which subsets of $K$ particles are interconnected via springs. \\
\textbf{VAR}:
The VAR model is a widely employed linear model to replicate dynamic behavior and analyze time series data. The VAR model is characterized by linear interactions. \\
\textbf{Netsim}: Netsim is a well-known dataset of quasi-real fMRI neural signals~\cite{smith2011network}. 

\subsection{Results}
\label{sec:mainresults}

  \subsubsection{Experiments with simulated datasets}
\label{sec:mainresults_simulated}

We show the inference accuracy for the three simulated datasets in Tab.~\ref{tab:main_resulsts}. We compared our DiffRI model with other state-of-the-art methods: NRI\cite{kipf2018neural}, NRI-MPM\cite{chen2021neural}, MI\cite{faes2011information, lizier2012multivariate}, and TE\cite{faes2011information,vicente2011transfer}. Training settings are described in Appx.~\ref{appen:experiment_sets_sec_5}. Model settings of baseline methods are described in Appx.~\ref{appx:baseline-methods}. In addition, we test DiffRI on scaled datasets. Results are shown in Appx.~\ref{appen:exp-scale}.

\begin{table*}[htbp]
\centering
\caption{\label{tab:main_resulsts} Network inference accuracy comparison across different datasets. For a fair comparison, we divided the upper five rows corresponding to the experiments without data augmentation and the lower one row corresponding to those with data augmentation.}
\begin{tabular}{lcllll}
\hline
\multirow{2}{*}{\textbf{Models}} & \multicolumn{2}{c}{\textbf{Kuramoto}}     & \multicolumn{2}{c}{\textbf{Spring}}                        & \multicolumn{1}{c}{\textbf{VAR}} \\
                                 & 5 nodes    & \multicolumn{1}{c}{10 nodes} & \multicolumn{1}{c}{5 nodes} & \multicolumn{1}{c}{10 nodes} & \multicolumn{1}{c}{5 nodes}      \\ \hline
NRI                              & 47.9 (4.0) & 52.8 (2.8)                   &  52.5 (2.4)                           & 51.2 (0.0)                             & 46.4 (8.9)                                 \\
NRI-MPM & 51.6 (1.8) & 52.8 (0.4) & 50.6 (0.8) & 49.7 (0.8) &  51.1 (3.8)\\
MI                              & 71.0 (10.8)         & 65.1 (6.3)                           & 70.0 (7.1)
& 59.1 (2.7)                         & 76.0 (4.2)             \\
TE                              & 77.0 (16.4)         & 70.4 (5.7)                         & 49.0 (5.4)                   & 50.0 (0.0)                         & \textbf{100.0 (0.0) }                                \\
DiffRI (proposed)                         & \textbf{92.6 (8.4)}                         & \textbf{95.3 (3.7)}                & \textbf{{97.5 (5.6)}}                & \textbf{98.0 (2.0)}                   & 88.1 (10.8)                      \\ \hline
%NRI (data-augmented) & 53.0 (13.0) & 60.1 (12.7)         & 94.5 (2.3)        & 82.5 (2.3) & 50.3 (1.4)    \\
NRI-MPM (data-augmented) & 74.4 (8.9) & 59.3 (5.3)  & \textbf{99.4 (0.6)} & 87.1 (8.0) & 51.2 (6.1) \\
\hline
\end{tabular}
\vspace{-0.0in}
\end{table*}
DiffRI shows high inference accuracy across three datasets. DiffRI achieves the best performance in 4 out of 5 data settings. Compared with information theory based methods MI and TE, we notice that TE gives the best performance in the VAR dataset, where the system is linear. However, TE and MI fail in the Kuramoto and 2D Spring datasets generated from nonlinear systems. Under fair experimental conditions (see Appx.~\ref{appen:datasets}, which gives minimal data to represent node features), DiffRI surpasses NRI and NRI-MPM in all five settings. However, with data augmentation, NRI-MPM performance improved and achieved the best performance in the Spring 5 nodes system. In addition to the performance in accuracy, we show the performance in Area Under the Receiver Operating Characteristic curve (AUROC) in Tab.~\ref{tab:auroc}.

\textbf{Discussions}: Here, we highlight the necessity of data augmentation for the previous benchmark. To demonstrate this, we conducted experiments by including additional data, such as particle velocity or first derivative, into the input data of NRI-MPM~\cite{chen2021neural}. See Appx.~\ref{appen:exp-aug} for details. The results are displayed in NRI-MPM (data-augmented). As shown in the below parts of Tab.~\ref{tab:main_resulsts}, inference accuracy improved compared with the corresponding above parts. Meanwhile, DiffRI can achieve similar or even superior performance without data augmentation. \\
In Tab.~\ref{tab:auroc}, the high AUROC value suggests that scores of edge existence and scores of edge non-existence are well distinguished intrinsically by DiffRI. The results show that the discrimination capability of DiffRI is high and insensitive to the threshold.

\begin{table}[]
\centering
\caption{\label{tab:auroc} AUROC results of DiffRI across different datasets.}
\begin{tabular}{ll}
\hline
\textbf{Datasets} & \textbf{AUROC}                   \\ \hline
Kuramoto-5     & 99.8 (0.4) \\
Kuramoto-10    & 98.1 (2.4)                        \\
Spring-5       & 99.0 (2.2)                        \\
Spring-10      & 99.7 (0.5)                        \\ \hline
\end{tabular}
\vspace{-0.2in}
\end{table}

\subsubsection{Experiments with Netsim datasets}
\label{sec:mainresults_netsim}
To further validate the effectiveness of the proposed method, we show the results of relational inference with DiffRI for three simulations within the Netsim datasets. The regularization term is omitted during training in experiments.

\begin{figure}[htbp]
  \centering
  \includegraphics[width=8cm,height=8.5cm]{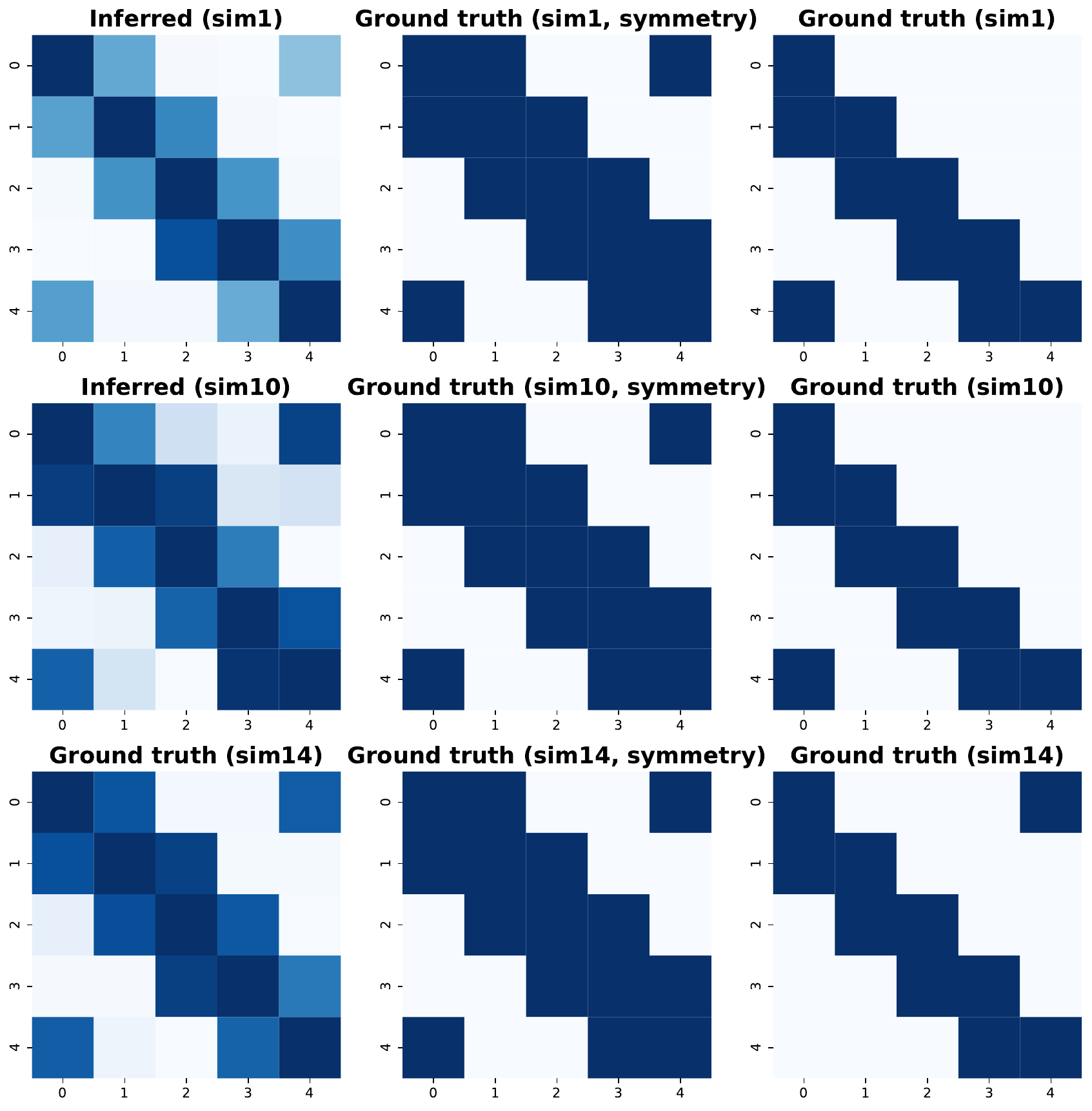}
  \caption{Illustration of DiffRI inference results and ground truth connection matrices on Netsim datasets. We show results on three simulations. The left column shows raw inference results. The middle column shows the symmetry ground truth connection matrix that ignores causal direction. The right column shows the ground truth connection matrix. Ground truth matrices are represented in a 0-1 binarized manner (blue: edge exists, white: edge does not exist). In the left column, a deeper color block indicates a larger value. }
\vspace{-0.1in}
  \label{fig:netsim_results}
\end{figure}

As demonstrated in the left and middle columns of Fig~\ref{fig:netsim_results}, DiffRI accurately identifies all pairs of nodes with relations under three different settings: $\textit{sim}_1$, $\textit{sim}_{10}$, and $\textit{sim}_{14}$ when directions are ignored. This indicates its robustness as a relational inference model. However, DiffRI struggles with determining causal directions. A comparison of the left and right columns reveals this confusion, suggesting that it may not be suitable for rigorous causal inference tasks yet.

\subsection{Ablation studies}
\label{sec:results-ablation studies}
We performed a set of ablation studies aimed at quantitatively assessing the impact of technical components.

First, three settings were used for ablation studies on regularization and the perturbation operator: DiffRI with regularization and perturbation operator (``DiffRI w/ Reg \& w/ PO "), DiffRI without regularization but with perturbation operator (``DiffRI w/o Reg \& w/ PO"), and DiffRI with regularization but without perturbation operation (``DiffRI w/ Reg \& w/o PO"). In ``DiffRI w/o Reg \& w/ PO", we let $\lambda_1=0$ in Eq.~(\ref{equ:loss}). The results for ``DiffRI w/ Reg \& w/ PO" are sourced from previous results in Section~\ref{sec:mainresults_simulated}. In the experiments with ``DiffRI w/ Reg \& w/o PO", the moving average operation was substituted with a Hadamard product, working as a 0/1 masking, formulated as $\textbf{x}^{k}_{\text{feat}} = f_{\text{LSTM}}(\widetilde{\mathbf{E}}^{k} \odot \mathbf{X}_{\text{temporal}})$. Experimental results are shown in Tab.~\ref{tab:ablationstudies}.
\begin{table*}[htbp]
\centering
\caption{\label{tab:ablationstudies} Ablation studies on DiffRI. Results are inference accuracy.}
\begin{adjustbox}{max width=\textwidth}
\begin{tabular}{l|clll}
\hline
\multicolumn{1}{c|}{\textbf{Model}} & \textbf{Kuramoto (5 nodes)} & \textbf{Kuramoto (10 nodes)} & \textbf{Spring (5 nodes)} & \textbf{Spring (10 nodes)} \\ \hline
DiffRI w/ Reg \& w/ PO                          &   92.6 (8.4)                          &   95.3 (3.7)                           &\textbf{97.5 (5.6)}  & \textbf{98.0 (2.0)}                        \\
DiffRI w/o Reg \& w/ PO                  & \textbf{100.0 (0.0)}                            & \textbf{97.5 (1.8)}                           & 78.2 (18.6)         & 85.1 (16.4)     \\
DiffRI w/ Reg \& w/o PO                     & 74.0 (15.6)                          &  83.1 (20.0)                            &    82.0 (24.9)    & 73.0 (19.9)                \\
 \hline
\end{tabular}
\end{adjustbox}
\vspace{-0.1in}
\end{table*}

\begin{table*}[htbp]
\centering
\caption{\label{tab:diffri-train} Ablation studies on DiffRI's training approaches. Results are inference accuracy. MR=Missing Ratio. Kura=Kuramoto}
\begin{adjustbox}{max width=\textwidth}
\begin{tabular}{lllllll}
\hline
\multicolumn{1}{c}{\multirow{2}{*}{\textbf{Models}}} & \multicolumn{2}{c}{\textbf{MR=0\%}} & \multicolumn{2}{c}{\textbf{MR=25\%}} & \multicolumn{2}{c}{\textbf{MR=50\%}} \\
\multicolumn{1}{c}{}                                 & Kura-5             & Kura-10           & Kura-5             & Kura-10            & Kura-5             & Kura-10            \\ \hline
DiffRI via prediction                        & 72.1 (14.8)            & 93.5 (4.6)            & 70.0 (12.2)            & 88.2 (5.7)             & 73.7 (11.8)            & 71.9 (11.4)            \\
DiffRI via imputation                        & \textbf{98.0 (4.5)}    & \textbf{97.1 (3.0)}   & \textbf{93.2 (8.8)}    & \textbf{95.8 (3.7)}    & \textbf{94.2 (5.3)}    & \textbf{93.2 (1.7)}    \\ \hline
\end{tabular}
\end{adjustbox}
\end{table*}

``DiffRI w/ Reg \& w/o PO" clearly performs worse than the other two baselines, which are equipped with perturbation operation. The presence of perturbation operations significantly enhances inference accuracy and reduces variance.

Upon comparison between ``DiffRI w/ Reg \& w/ PO" and ``DiffRI w/o Reg \& w/ PO", it is evident that neither configuration consistently outperforms the other. These findings imply that regularization utility may be considered auxiliary, depending on the data.

Second, DiffRI utilizes imputation instead of prediction for relation inference. Here, we show that imputation training performs better than the prediction baseline. We train DiffRI via prediction by modifying the target mask to the prediction mode (predict consecutive future time steps of the target time series). The percentages of conditional data and target data are the same as those in imputation training. Other hyper-parameter values are maintained the same. Results are summarized in Tab.~\ref{tab:diffri-train}. From Tab.~\ref{tab:diffri-train}, the imputation training approach consistently performs better. Meanwhile, as the missing ratio increases, the accuracy of the model trained via imputation does not suffer so much. In contrast, the accuracy declines significantly for the prediction baseline.

\section{Conclusion}
In this work, we proposed DiffRI, a diffusion generative-based model for relational inference from time series. DiffRI extracts latent relational information by imputing time series. Experiments on three interacting systems show that DiffRI is highly competent in inferring underlying relations. Moreover, we found that DiffRI can correctly infer relations in quasi-real brain signal datasets with various realistic settings. In addition, ablations studies confirmed the effectiveness of the perturbation operator and imputation training in DiffRI.

\bibliography{reference}

%%%%%%%%%%%%%%%%%%%%%%%%%%%%%%%%%%%%%%%%%%%%%%%%%%%%%%%%%%%%%%%%%%%%%%%%%%%%%%%
%%%%%%%%%%%%%%%%%%%%%%%%%%%%%%%%%%%%%%%%%%%%%%%%%%%%%%%%%%%%%%%%%%%%%%%%%%%%%%%
% APPENDIX
%%%%%%%%%%%%%%%%%%%%%%%%%%%%%%%%%%%%%%%%%%%%%%%%%%%%%%%%%%%%%%%%%%%%%%%%%%%%%%%
%%%%%%%%%%%%%%%%%%%%%%%%%%%%%%%%%%%%%%%%%%%%%%%%%%%%%%%%%%%%%%%%%%%%%%%%%%%%%%%
\newpage
\appendix
\onecolumn
\section{Theoretical analyses}
\subsection{Equivalence between denoising loss in Eq.~(\ref{equ:loss}) and KL divergence}
\label{appen:kldiv}
We show that noise approximation loss in Eq.~(\ref{equ:loss}) is equivalent to the KL divergence between the ground truth reverse probability distribution $q(\mathbf{x}_{t-1}^k|\mathbf{x}_t^k)$ and the approximation probability distribution $p_\theta(\mathbf{x}_{t-1}^k|\mathbf{x}_t^k,\mathbf{X}^{\text{co}}, \mathbf{E}^k)$ of the reverse process. As defined in Sec.~\ref{sec:framework}, $k$ is the index of the target node, $t$ denotes the diffusion step, and $\mathbf{x}_t^k$ ($1\le t \le T$) represent the latent variables in the diffusion process regarding the imputation target $\mathbf{x}_0^k$ in time series at node $k$. During derivation, we will also show that minimizing the noise approximation loss is equivalent to minimizing the estimation error of original data $\mathbf{x}_0^k$. Our derivation is mainly based on a previous study~\cite{luo2022understanding}. For simplicity, we omit the expectation mark $\mathbb{E}$ in the following. The KL divergence is shown as follows:
\begin{equation}
\min_\theta D_{\text{KL}}(q(\mathbf{x}_{t-1}^k|\mathbf{x}^k_t)||p_\theta(\mathbf{x}_{t-1}^k|\mathbf{x}_t^k,\mathbf{X}^{\text{co}}, \mathbf{E}^k)).     
\end{equation}
Both distributions are Gaussian. Therefore, we can write the KL divergence term as:
\begin{align}
&\min_\theta D_{\text{KL}}(\mathcal{N}(\mathbf{x}_{t-1}^k;\mu^k_q(\mathbf{x}^k_t),\Sigma^k_q(t))||\mathcal{N}_\theta(\mathbf{x}_{t-1}^k;\mu_{\theta}(\mathbf{x}^{k}_{t},\mathbf{X}^{\text{co}}, \mathbf{E}^k,t),\Sigma^k_q(t))  \nonumber \\
=&\min_\theta \frac{1}{2} [\log\frac{|\Sigma^k_q(t)|}{|\Sigma^k_q(t)|} ]-d+\text{tr}(\Sigma^k_q(t)^{-1}\Sigma^k_q(t)) + (\mu^k_{\theta}-\mu^k_q)^T\Sigma^k_q(t)^{-1}(\mu_{\theta}-\mu_q).
\end{align}
Following the definition, $\Sigma^k_q(t)=\sigma_q^2(t)\mathbf{I}$ is diagonal and does not contain trainable parameters. Hence we have
\begin{align}
&\min_\theta \frac{1}{2}[(\mu^k_{\theta}-\mu^k_q)^T\Sigma^k_q(t)^{-1}(\mu^k_{\theta}-\mu^k_q) \nonumber \\
=&\min_\theta \frac{1}{2\sigma^2_q(t)}[||\mu^k_\theta-\mu^k_q||_2^2].
\end{align}
Then, we first parameterized $\mu$ with $\mathbf{x}_t$ and $\mathbf{x}_0$. Let $\bar{\beta}_t = \prod_{i=1}^{i=t} (1-\beta_i)$. Following previous literature~\cite{luo2022understanding}, the mean function can be written as follows:
\begin{align}
&\mu^k_q(\mathbf{x}_t^k,\mathbf{x}_0^k)=\frac{\sqrt{1-\beta_t}(1-\bar{\beta}_{t-1})\mathbf{x}_t^k+\sqrt{\bar{\beta}_{t-1}}\beta_t\mathbf{x}_0^k}{1-\bar{\beta}_t},\\
&\mu^k_\theta(\mathbf{x}^{k}_{t}, \mathbf{X}^{\text{co}}, \mathbf{E}^k,t) = \frac{\sqrt{1-\beta_t}(1-\bar{\beta}_{t-1})\mathbf{x}_t^k+\sqrt{\bar{\beta}_{t-1}}\beta_t\hat{\mathbf{x}}_\theta^k(\mathbf{x}^{k}_{t},\mathbf{X}^{\text{co}}, \mathbf{E}^k,t)}{1-\bar{\beta}_t}.
\end{align}
Therefore, we can rewrite the mean difference loss as follows:
\begin{align}
&\min_\theta \frac{1}{2\sigma_q(t)^2}[||\mu^k_\theta-\mu^k_q||_2^2] \nonumber\\
=&\min_\theta \frac{1}{2\sigma_q(t)^2} \frac{\bar{\beta}_{t-1}\beta^2_t}{(1-\bar{\beta}_t)^2}[||\hat{\mathbf{x}}_\theta^k(\mathbf{x}^{k}_{t}, \mathbf{X}^{\text{co}}, \mathbf{E}^k,t)-\mathbf{x}_0^k||_2^2].
\end{align}
This highlights that \textit{minimizing KL divergence is equivalent to minimizing the estimation error of the original data}. Meanwhile, following the forward process definition, we can parameterize $\mathbf{x}_0$ with $\mathbf{x}_t$ and $\epsilon^k$ as follows: 
\begin{align}
& \mathbf{x}_0^k = \frac{\mathbf{x}_t^k-\sqrt{1-\bar{\beta}_t}\epsilon^k}{\sqrt{\bar{\beta}_t}}.
\end{align}
We substitute the above expression into the expression of mean functions $\mu_\theta^k$ and $\mu_q^k$. Then, the mean difference loss can be rewritten as follows:
\begin{align}
&\min_\theta \frac{1}{2\sigma_q(t)^2}[||\mu^k_\theta-\mu^k_q||_2^2] \nonumber\\
=&\min_\theta \frac{1}{2\sigma_q(t)^2} \frac{\beta^2_t}{(1-\bar{\beta}_t)(1-\beta_t)}[||\epsilon^k_\theta(\mathbf{x}_t^k,\mathbf{X}^{\text{co}}, \mathbf{E}^k,t)-\epsilon^k||_2^2].
\end{align}
We obtain the adopted loss function in Eq.~(\ref{equ:loss}).  This emphasizes that \textit{minimizing KL divergence is equivalent to minimizing the difference between actual and estimated noise}.

\subsection{Theoretical interpretation of DiffRI with the regularization term}
\label{appen:theo_inter}
We provide a theoretical interpretation of DiffRI. In Sec.~\ref{appen:kldiv}, we show that minimizing denoising loss is equivalent to minimizing the KL divergence $D_{\text{KL}}(\mathcal{N}(\mathbf{x}_{t-1}^k;\mu^k_q(\mathbf{x}^k_t),\Sigma^k_q(t))||\mathcal{N}_\theta(\mathbf{x}_{t-1}^k;\mu_{\theta}(\mathbf{X}^{\text{co}}, \mathbf{E}^k,t),\Sigma^k_q(t))$. We further show that, under certain approximations, minimizing the objective function in Eq.~(\ref{equ:loss}) is equivalent to minimizing the cross entropy with constraints. 

First, considering the equivalence between denoising loss and KL divergence, we rewrite Eq.~(\ref{equ:loss}) as the following optimization problem:
\begin{align}
\begin{split}
    \min_\theta &~D_{\text{KL}}(q(\mathbf{x}_{t-1}^k|\mathbf{x}^k_t)||p_\theta(\mathbf{x}_{t-1}^k|\mathbf{x}_t^k,\mathbf{X}^{\text{co}}, \mathbf{E}^k)) \\
    %\label{eq:appen_kl} \\
    \text{s.t. } & \left\lVert\frac{\sum_{i,i\neq k}{\mathbf{E}^{k}}[i]}{K-1}-\rho \right\rVert = 0. \label{eq:appen_constraint}
\end{split}
\end{align}
Then, we consider how to simplify $p_\theta(\mathbf{x}_{t-1}^k|\mathbf{x}_t^k,\mathbf{X}^{\text{co}}, \mathbf{E}^k,t)$. Notice that the feature interaction layer with the perturbation operator, described in Sec.~\ref{sec:feature_inter}, is the only component in DiffRI that allows information from different time series to interact. In our implementation, we let non-interacting time series be moving averaged. However, it is difficult to model the moving average in probabilistic expression. For derivation purposes, we approximate this by considering the case that non-interacting time series are totally removed from the approximated reverse probability $p_\theta$. We rewrite the objective function in Eq.~(\ref{eq:appen_constraint}) as follows:
\begin{align}
    &D_{\text{KL}}(q(\mathbf{x}_{t-1}^k|\mathbf{x}^k_t)||p_\theta(\mathbf{x}_{t-1}^k|\mathbf{x}_t^k,\{\mathbf{x}_0^i;\mathbf{e}^{k,i}[0]=1\})) \nonumber\\
    =&\int q(\mathbf{x}_{t-1}^k|\mathbf{x}^k_t) \log\frac{q(\mathbf{x}_{t-1}^k|\mathbf{x}^k_t)}{p_\theta(\mathbf{x}_{t-1}^k|\mathbf{x}_t^k,\{\mathbf{x}_0^i;\mathbf{e}^{k,i}[0]=1\})} \nonumber\\
    =&\int  q(\mathbf{x}_{t-1}^k|\mathbf{x}^k_t)\log q(\mathbf{x}_{t-1}^k|\mathbf{x}^k_t) - q(\mathbf{x}_{t-1}^k|\mathbf{x}^k_t)\log p_\theta(\mathbf{x}_{t-1}^k|\mathbf{x}_t^k,\{\mathbf{x}_0^i;\mathbf{e}^{k,i}[0]=1\}). \label{eq:app_klexpand}
\end{align}
Notice that the term $q(\mathbf{x}_{t-1}^k|\mathbf{x}^k_t)$ does not contain any trainable parameter. Therefore, we can rewrite the optimization problem given by Eq.~(\ref{eq:appen_constraint}) as follows:
\begin{align}
\begin{split}
    \min_\theta &~\int  - q(\mathbf{x}_{t-1}^k|\mathbf{x}^k_t)\log p_\theta(\mathbf{x}_{t-1}^k|\mathbf{x}_t^k,\{\mathbf{x}_0^i|\mathbf{e}^{k,i}[0]=1\}) \label{eq:app_crossE} \\
    \text{s.t. } & \left\lVert\frac{\sum_{i,i\neq k}{\mathbf{E}^{k}}[i]}{K-1}-\rho \right\rVert = 0. 
\end{split}
\end{align}
Eq.~(\ref{eq:app_crossE}) is exactly the cross entropy between ground truth reverse probability $q(\mathbf{x}_{t-1}^k|\mathbf{x}^k_t)$ and estimated reverse probability $p_\theta(\mathbf{x}_{t-1}^k|\mathbf{x}_t^k,\{\mathbf{x}_0^i|\mathbf{e}^{k,i}[0]=1\})$. Note that cross-entropy quantifies the expected message length to describe ground truth distribution $q(\mathbf{x}_{t-1}^k|\mathbf{x}^k_t)$ using the approximated distribution $p_\theta(\mathbf{x}_{t-1}^k|\mathbf{x}_t^k,\{\mathbf{x}_0^i|\mathbf{e}^{k,i}[0]=1\})$~\cite{bishop2006pattern}. With all of the above, we conclude that DiffRI (w/ an optional regularization) \textit{seeks a fixed number of time series that can minimize expected message length for $p_\theta$ to describe ground truth reverse probability $q(\mathbf{x}_{t-1}^k|\mathbf{x}_{t}^k)$.} 

\section{More details on DiffRI implementations}
\subsection{Neural network implementation of the noise approximation model}
\label{appenx:Neural network implementation of the noise approximation function}

We adopted the diffusion noise approximation neural network from CSDI's code, as seen in Fig.~6 of \cite{tashiro2021csdi}. The primary components of the CSDI network architecture include a temporal transformer layer, a feature transformer layer, and several 1D convolution layers. The feature transformer equates to a fully connected GNN (which does not necessarily adhere to relations inferred by the edge prediction module), which conflicts with the intentions of DiffRI. Therefore, we introduced a feature interaction layer (an LSTM or MLP module as the backbone) with a perturbation operation to replace the feature transformer layer, as detailed in Sec.~\ref{sec:feature_inter}. Moreover, we let the temporal transformer respect causal by adding a causal attention mask.

Regarding hyperparameter settings, we use the parameter values in Tab.~\ref{tab:diffnri_implement} for all DiffRI experiments in our study.

\begin{table}[htbp]
\centering
\caption{\label{tab:diffnri_implement} Implementation settings for noise approximation model in DiffRI.}
\begin{tabular}{lc}
\hline
\textbf{Parameter}      & \textbf{Value} \\ \hline
Residual layers         & 1 (for Netsim) / 2              \\
Channels                & 64    \\
Number of hidden units $H$      & 64   \\
Attention heads         & 1 (for Netsim) / 4              \\
Diffusion embedding dim & 128            \\
Diffusion steps         & 50             \\
Beta start/end         & 0.0001/0.5     \\
\hline
\end{tabular}
\end{table}

The output of the noise approximation network, $\epsilon_\theta^k$, is an estimated noise of the imputation target. We train the diffusion generative model with the denoising loss (the difference between estimated noise and actual noise) and an optional regularization loss.

\subsection{Masking strategy}
\label{appen:masking_strategy}
In DiffRI, we divide time series data into target and conditional segments using a masking technique. We show an algorithm to generate two masks for the purpose in Algo.~\ref{alg:generate_mask}.

For each sample in a mini-batch, we randomly pick a target time series, denoted by $A^{\text{target}}[i]$ in Algo.\ref{alg:generate_mask}. We set the default masking ratio $r$ at 0.5, meaning that we randomly mask half of the data within the target time series to form imputation targets, denoted by $M^{\text{target}}$ in Algo.~\ref{alg:generate_mask}. The remaining observed data constitutes conditional data, selected by $M^{\text{conditional}}$. Both masks have dimensions $B\times K\times L$, where $B$ is the batch size, $K$ represents the number of components in the system, and $L$ stands for time length. To obtain imputation targets $\mathbf{x}^k_0$ and conditional data $\mathbf{X}^{\text{co}}$, we multiply $M^{\text{target}}$ and $M^{\text{conditional}}$ by input data respectively on an element-wise basis.

For DiffRI trained via prediction, we simply modify the $M^{\text{target}}$ mask. Consecutive future points in  $M^{\text{target}}$ were set to be 1s, while corresponding $M^{\text{conditional}}$ parts were set to be 0s.

%\begin{lstlisting}[language=Python, caption=Generate mask for relational inference]
% def get_inter_mask(self, observed_mask):
%     B, K, L = observed_mask.shape
%     cond_mask = torch.zeros_like(observed_mask) 
%     target_mask = torch.zeros_like(observed_mask) 
%     rand_m = torch.rand_like(observed_mask) * observed_mask
%     target_list = torch.zeros(B)
%     for i in range(B):
%         target_k = random.sample(range(K), 1)
%         target_list[i] = target_k[0]
%         for j in range(K):
%             if j == target_k[0]:
%                 sample_ratio = 0.5
%             else:
%                 sample_ratio = 0
%                 continue
%             num_observed = sum(observed_mask[i,j,:])
%             num_masked = (num_observed * sample_ratio).round()
%             rand_m[i,j,:][rand_m[i,j].topk(int(num_masked)).indices] = -1
%             cond_mask[i,j,:][rand_m[i,j] > 0] = 1
%         target_mask[i,target_k[0],:][rand_m[i,target_k[0],:]==-1] = 1
%     return cond_mask==1, target_mask==1, target_list
% \end{lstlisting}

\begin{algorithm}[tb]
   \caption{Mask generation for determining conditional parts and target parts}
   \label{alg:generate_mask}
\begin{algorithmic}
   \STATE {\bfseries Input:} Batch size $B$, Number of components $K$, Time steps length $L$, Observed mask $M^{\text{observed}}$
   \STATE {\bfseries Output:} $M^{\text{conditional}}, M^{\text{target}}$
   \STATE Initialize two masks $M^{\text{conditional}}, M^{\text{target}}$ with shape $B\times K\times L$, a list $A^{\text{target}}$ with length $K$, a sample ratio $r$, an integer $\text{num}_{\text{masked}}$, and a random matrix $M^{\text{rand}}$ with shape $B\times K\times L$ and elements follow i.i.d uniform distribution on the interval [0,1).
   \FOR{$i=1$ {\bfseries to} $B$}
        \STATE \text{Let} $A^{\text{target}}[i] = \text{a random integer between 1 and K}$

        \STATE $r = 0.5$
        \STATE $\text{num}_{\text{masked}} = \text{SUM}(M^{\text{observed}}[i,A^{\text{target}}[i]]) * r$
        \STATE $M^{\text{rand}}[i,A^{\text{target}}[i],M^{\text{rand}}[i,j].\text{topk}(\text{num}_{\text{masked}}).\text{indices}] = -1$
        \STATE $M^{\text{conditional}}[i,A^{\text{target}}[i],:][M^{\text{rand}}[i,j]>0]=1$
        \STATE $M^{\text{target}}[i,A^{\text{target}}[i],:][M^{\text{rand}}[i,A^{\text{target}}[i],:]=-1]=1$
   \ENDFOR
\end{algorithmic}
\end{algorithm}

\section{Baseline methods}
\label{appx:baseline-methods}
In our study, we consider the following well-known baseline models: Neural Relational Inference (NRI), Neural Relational Inference with Message Passing Mechanism (NRI-MPM), Mutual Information (MI), and Transfer Entropy (TE). At present, we have not located the code for some state-of-the-art methods, e.g., ~\cite{wang2023effective}. 
\begin{itemize}
  \item NRI \cite{kipf2018neural}: a representative prediction-based VAE model for unsupervised relational inference.
  \item NRI-MPM \cite{chen2021neural}: an improved version of NRI with efficient message passing and structural prior.
  \item MI \cite{faes2011information, lizier2012multivariate}: a multivariate mutual information-based method for network structure inference, using Information Dynamics Toolkit xl (IDTxl).
  \item TE \cite{faes2011information,vicente2011transfer}: a multivariate transfer entropy-based method for network structure inference, using IDTxl.
\end{itemize}

\textbf{NRI}: We used the public code\footnote{https://github.com/ethanfetaya/NRI} for implementations. The main settings of model components and hyperparameters are shown in the Tab.~\ref{tab:nri_baseline}. For all the other hyperparameters, we followed the default setting in the public train script. We run every experiment for 500 epochs. Through tuning hyper-parameters (changing encoder and decoder hidden units ranging from 64,128 and 256; dropout probability ranging from 0, 0.5, skip-first and no skip-first), we found that the provided settings give better performance. We ensure that the models reach convergence during training.

\begin{table}[htbp]
\centering
\caption{\label{tab:nri_baseline} Model architecture settings and hyperparameters for NRI.}
\begin{tabular}{lc}
\hline
\textbf{Component/Parameter}      & \textbf{Model/Value} \\ \hline
Encoder        & CNN            \\
Decoder                & MLP    \\
Encoder hidden units         & 256             \\
Decoder hidden units         & 256    \\
Prediction steps         & 10              \\
Epoch & 500 \\
Batch size & 16 \\
Skip first & True \\\hline
\end{tabular}
\end{table}

\textbf{NRI-MPM}: NRI-MPM introduces efficient message-passing mechanisms to the graph neural networks with structural prior knowledge to the previous NRI model~\cite{chen2021neural}. NRI-MPM has achieved better performance compared with other models like NRI, SUGAR, and ModularMeta. Therefore, we selected NRI-MPM as one of the baselines. The main settings of model components and hyperparameters are shown in Tab.~\ref{tab:nrimpm_baseline}. We used the public code\footnote{https://github.com/hilbert9221/NRI-MPM} for implementation. We noticed that this code includes a bug in the part of KL divergence loss. Therefore, we corrected it in our experiments. We run the NRI-MPM model for Kuramoto and Spring datasets following the author's recommended public README and configuration. Through tuning hyper-parameters, we found that the provided settings give better performance. We ensure that the models reach convergence during training.

\begin{table}[htbp]
\centering
\caption{\label{tab:nrimpm_baseline} Model architecture settings and hyperparameters for NRI-MPM.}
\begin{tabular}{lc}
\hline
\textbf{Component/Parameter}      & \textbf{Model/Value} \\ \hline
Encoder        & RNN            \\
Decoder                & RNN    \\
Encoder hidden units         & 256             \\
Decoder hidden units         & 256    \\
Reduce network & CNN/MLP\\
Epoch & 500 \\
Batch size & 64 \\
\hline
\end{tabular}
\end{table}

\textbf{MI and TE}:
We used the public IDTxl package for MI and TE experiments, which offers methods for inferring network structure based on multivariate mutual information and transfer entropy\footnote{https://github.com/pwollstadt/IDTxl}~\cite{wollstadt2019idtxl}. For more details about these algorithms, refer to the official tutorial\footnote{https://github.com/pwollstadt/IDTxl/wiki/Theoretical-Introduction}. The main settings of model components and hyperparameters listed in Tab.~\ref{tab:idtxl_para} are used in both MI and TE benchmark experiments, with all the other parameters left at their default settings.

\begin{table}[htbp]
\centering
\caption{\label{tab:idtxl_para} Parameters for MI and TE baseline experiments.}
\begin{tabular}{lc}
\hline
\multicolumn{1}{c}{\textbf{Component/Parameter}} & \textbf{Model/Value}          \\ \hline
cmi\_estimator                & JidtKraskovCMI \\
n\_perm\_max\_stat            & 100            \\
n\_perm\_min\_stat            & 100            \\
n\_perm\_omnibus              & 100            \\
n\_perm\_max\_seq             & 100            \\
max\_lag\_sources             & 5              \\
min\_lag\_sources             & 1            \\ \hline
\end{tabular}
\end{table}

\section{Details on datasets}
\label{appen:datasets}

\begin{figure}[htbp]
  \centering
  \includegraphics[width=8.4cm,height=3.3cm]{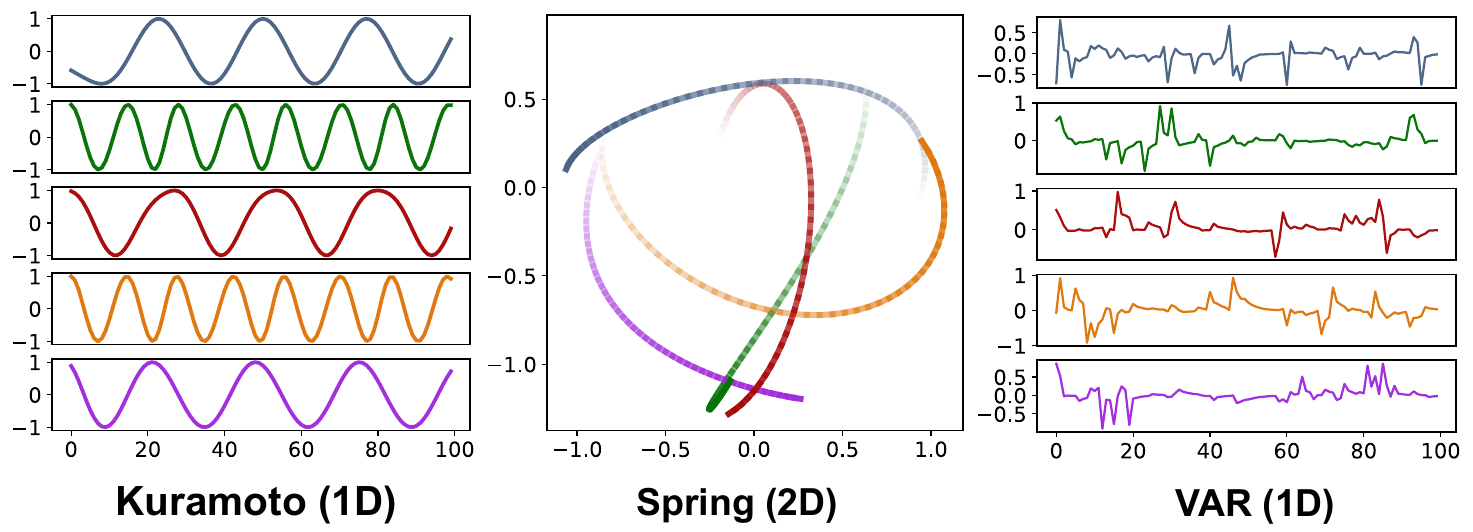}
  \caption{Example of simulated datasets used in our experiments: Kuramoto model (left), Spring (middle), and VAR (right). }
  \label{fig:data_example}
  \vspace{-0.2in}
\end{figure}

In this section, we give a detailed introduction of the four datasets we considered. Fig.~\ref{fig:data_example} showcases examples of simulated data.

\paragraph{Kuramoto dataset}
In our study, we employed a generalized version of the Kuramoto model, where the connectivity is not full, as follows:
\begin{equation}
\Dot{\theta_i}(t) = \omega_i + \frac{C}{K_i}\sum_{j=1}^{K}w_{ij} \sin(\theta_j(t)-\theta_i(t)),\quad i=1,...,K,
\label{equ:kuramoto}
\end{equation}
where $\theta_i$ is the phase of oscillator node $i$, $K$ is the number of nodes, $\mathbf{W}=( w_{ij})$ is a symmetric coupling matrix, and $C$ is the coupling strength. The natural frequency $\omega_i$ is sampled from a uniform distribution on (2,10). The parameter $K_i$ is the degree of oscillator node $i$. Different from previous works~\cite{kipf2018neural, zhang2019general, chen2021neural}, which leverage the information on frequencies $\omega_i$ or first time derivative $d\theta_i(t)/dt$ to augment data, we only used the values of $\text{sin}(\theta_i(t))$ as observables representing the dynamical behavior of oscillator $i$. We simulated the dataset using publicly available code\footnote{https://github.com/fabridamicelli/kuramoto}. %(except experiments with NRI-MPM).

The Kuramoto dataset contains a set of multivariable time series data representing the time courses of $\sin{\theta}_i(t)$ in Eq.~(\ref{equ:kuramoto}). For both 5-node system and 10-node system, we generate a training dataset with 500 samples, a validation set with 100 samples, and a test set with 100 samples. Unless otherwise stated, each time series is produced using a total of 100 time steps (with total time length $T=5$ and time step $dt=0.05$). Using five random number seeds, we generated five distinct datasets where the samples within each dataset share an identical underlying network (we formulate the studied problem as a transductive learning problem). Our model was trained on each dataset, and the results were averaged out.

\paragraph{Spring dataset}
The simulation of the Spring model produces 2D trajectories of these particles. Interactions between connected particles adhere to Hooke's law. That is, the acceleration of each particle is contingent upon the separation distance between the particle and its connected neighbors. We generated 2D spring datasets using the code publicly available in the NRI GitHub repository\footnote{https://github.com/ethanfetaya/NRI/tree/master/data}. However, unlike the NRI model, which leverages both trajectory and velocity inputs, we only use 2D trajectory coordinates as model inputs unless otherwise noted.

The Spring dataset includes multivariate time series data corresponding to trajectories of particles (in 2D space) coupled by springs. We generated a training set with 500 samples (validation set, test set: 100 samples) for the 5-node system and another with 1500 samples (validation set, test set: 300 samples) for the 10-node system. For DiffRI, trajectories in the two coordinates were concatenated to form a time series. Unless specified differently, time series were generated using a total of 49 steps (with total time length $T=5000$ and sampling frequency of 100). By utilizing five distinct random number seeds, we produced five distinct spring datasets where samples in each dataset share an identical network structure, but these structures differ across datasets. Our model was trained on each separate dataset, and the results were averaged out.

\paragraph{VAR dataset}
In our study, the VAR model used to generate data is expressed as follows:
\begin{equation}
y_i(t) = \sum_{p=1}^{P}\mathbf{A} \mathbf{y}_{t-p},\quad i=1,...,K,
\label{equ:var}
\end{equation}
where $\mathbf{A}$ is a symmetric matrix whose $(i,j)$ element is given by a random positive or negative value if node $i$ and $j$ are interacting and zero otherwise. To mimic real-world scenarios, we randomly let data $y_i(t)$ become noisy.

The VAR dataset consists of multivariate time series where the $i$-th row corresponds to $y_i(t)$ in Eq.~(\ref{equ:var}). The VAR dataset contains 500 training samples, 100 validation samples, and 100 test samples. Unless otherwise stated, we set the total time length at $T=100$. % and the VAR model is the first order shown as the following code list. 
% \begin{lstlisting}[language=Python, caption=VAR data generation]
%     for t in range(1, T):
%         for n in range(N):
%             rate = np.random.rand()
%             if rate < 0.1:
%                 y[t, n] = np.random.uniform(size=(1), low=-1, high=1)
%             else:
%                 y[t, n] = np.matmul(ar_coef[n,:], y[t - 1]) 
% \end{lstlisting}

\paragraph{Netsim dataset}
Netsim is a well-known dataset of quasi-real fMRI neural signals~\cite{smith2011network}. There are several factors that make this dataset challenging. Firstly, the connection strengths within each simulation are not constant and vary across subjects. Secondly, unobservable external inputs influence the system's behavior, simulating inputs from other brain regions. Lastly, the dataset includes both neural noise and measurement noise. For further details, refer to the paper~\cite{smith2011network}. Three simulations were chosen for our experiments, $\textit{sim}_1$, $\textit{sim}_{10}$ and $\textit{sim}_{14}$. In addition to the shared properties, $\textit{sim}_{10}$ contains global mean confound and $\textit{sim}_{14}$ has cyclic connections.

The Netsim dataset comprises 28 simulations, three of which were chosen for our experiments in Sec.~\ref{sec:mainresults_netsim}. Each simulation includes 50 samples, each with five time series of length 200. We did not separate the samples into training and test datasets because the dataset size is small, and DiffRI does not have access to ground truth relations during training.

%\textcolor{red}{Second, we focus on transductive relational inference tasks. Experimental results in NRI and NRI-MPM papers are from inductive relational inference tasks. Success in inductive learning requires a large number of samples and data augmentation techniques. When data samples are limited, and data augmentation is removed, performance drops for NRI and NRI-MPM. 

Here, we elaborate on the reason for focusing on small datasets. Large datasets are not realistic in many scenarios. For example, in healthcare, collecting large datasets is often impossible since only a certain number of medical institutions have enough cases. Although some techniques, such as Federated Learning, seem practical in real scenarios, relevant laws for privacy protection~\cite{floca2014challenges} and the realistic difficulty of many healthcare data systems to provide consistent data~\cite{kakulapati2023analysis} make local learning with a relatively small amount of data still be the first choice for the current situation. This is why we've chosen to concentrate on transductive relational inference tasks involving small datasets. It is worth noticing that experimental results in NRI and NRI-MPM papers~\cite{kipf2018neural,chen2021neural} are from inductive relational inference tasks. Success in inductive learning requires large samples and data augmentation. When data samples are limited, and data augmentation is removed, performance decreases for NRI-MPM, as shown in Tab.~\ref{tab:main_resulsts}. 

\section{Details of experiments and additional experiments}
\subsection{Experiment settings for  Sections~\ref{sec:mainresults_simulated}} 
\label{appen:experiment_sets_sec_5}
All DiffRI experiments were run using the Adam optimizer with an initial learning rate of 0.0005, decayed by a factor of 0.5 at several stages. Unless otherwise noted, we trained the model with a batch size equal to 16. All experiments are conducted on a single NVIDIA Tesla V100-SXM2 graphic card within an NVIDIA DGX-1 server. We list the other parameter settings for DiffRI in Tab.~\ref{tab:DiffRI-main-results-para}. Note that epoch equals 2000 is the upbound epochs number. In experiments, training ends earlier if validation loss does not decrease.

% \begin{table}[htbp]
% \centering
% \begin{tabular}{lc}
% \hline
% \textbf{Parameter}      & \textbf{Value} \\ \hline
% Epoch & 2000 \\ 
% Number of diffusion step $T$ & 50\\
% Adam weight decay    & 1e-6\\
% Validation epoch interval & 25 \\
% Early stop patience& 5 \\
% Feature interaction layer & LSTM (simulated), MLP (Netsim)    \\
% \hline
% \end{tabular}
% \end{table}

% \usepackage{multirow}
\begin{table}[htbp]
\centering
\caption{\label{tab:DiffRI-main-results-para} Model architecture settings and hyperparameters for DiffRI.}
\begin{tabular}{l|cccc}
\hline
\multicolumn{1}{c|}{\multirow{2}{*}{\textbf{Parameters}}} & \multicolumn{4}{c}{\textbf{Values}}                                                                              \\ \cline{2-5} 
\multicolumn{1}{c|}{}                                     & \multicolumn{1}{l}{Kuramoto} & \multicolumn{1}{l}{Spring} & \multicolumn{1}{l}{VAR} & \multicolumn{1}{l}{Netsim} \\ \hline
Feature interaction layer                                 & LSTM                         & LSTM                       & LSTM                    & MLP                        \\
$\lambda_1$                                               & 0.01                          & 0.01                       & 0.01                        & 0.00                  \\
Epoch                                                     & 2000                         & 2000                       & 2000                    & 2000                       \\
Number of diffusion step $T$                              & 50                           & 50                         & 50                      & 50                         \\
Early stop patience                                       & 10                            & 10                         & 5                       & 5                          \\
Adam weight decay                                         & 1e-6                         & 1e-6                       & 1e-6                    & 1e-6                       \\
Validation epoch interval                                 & 25                           & 25                         & 25                      & 25                         \\ \hline
\end{tabular}
\end{table}

\subsection{Baseline experiments with augmented data}
\label{appen:exp-aug}
% Baseline results detailed in Sec.~\ref{sec:experiments} don't match up to those reported in the original paper. In Sec.~\ref{sec:experiments}, we evaluated baseline models using the identical dataset employed for DiffRI (see Appx.~\ref{appen:datasets}). We highlight here that this discrepancy in performance stems from data augmentation practices. To illustrate this, we repeated some experiments incorporating particle velocity or first derivative into the input data. Other settings of datasets are kept unchanged. Results are averaged across several seeds. We show results in the Tab.~\ref{tab:aug_exps} below.

% \begin{table}[htbp]
% \centering
% \begin{tabular}{lclll}
% \hline
% \multirow{2}{*}{\textbf{Models}} & \multicolumn{2}{c}{\textbf{Kuramoto}}               & \multicolumn{2}{c}{\textbf{Spring}}                        \\
%                                  & 5 nodes              & \multicolumn{1}{c}{10 nodes} & \multicolumn{1}{c}{5 nodes} & \multicolumn{1}{c}{10 nodes} \\ \hline
% NRI                              & 47.9 (4.0)           & 49.1 (1.4)                   & 52.5 (2.4)                  & 51.2 (0.0)                   \\
% NRI (data augmented)             & \textbf{53.0 (13.0)} & \textbf{60.1 (12.7)}         & \textbf{86.4 (5.9)}         & \textbf{82.5 (2.3)}          \\ \hline
% \end{tabular}
% \caption{\label{tab:aug_exps} Comparing model performance under different input data settings.}
% \end{table}

In the experiment with the NRI-MPM benchmark, we augmented the Kuramoto datasets with oscillators' intrinsic frequency $\omega_i$, the time course of the first time derivatives $d\theta_i(t)/dt$, and $\theta_i(t)$, as in \cite{chen2021neural}. For the Spring dataset, we supplemented input data with particle velocities $(v_x,v_y)$. We enhanced the VAR datasets with the first derivatives $dy_i(t)/dt$. As shown in Tab.~\ref{tab:main_resulsts}, across all five datasets, the NRI-MPM model performs better when augmented with additional data compared to its non-augmented counterpart. The performance gap suggests that augmenting input data can notably boost the performance of the benchmark model, a factor often neglected in prior studies. In contrast, DiffRI achieves comparable or superior results without data augmentation. 

\subsection{Experiments with scaled datasets}
\label{appen:exp-scale}
To check scalability, we tested DiffRI for a larger interacting system. We introduce the settings for experiments using a scaled dataset and their results. All experiments are conducted on a single NVIDIA Tesla V100-SXM2 graphic card within an NVIDIA DGX-1 server.

We generated a set of multivariate time series data by numerically simulating a 25-node Kuramoto model and divided them into 1000 training samples, 200 validation samples, and 200 testing samples. The regularization term is omitted in model training. We found that DiffRI is highly accurate in inferring interactions in the Kuramoto system. Averaged results over 5 trials with different random number seeds indicate \textit{an accuracy of $96.5\%$ with $1.7\%$ std}.

\begin{figure*}[htbp]
  \centering
  \includegraphics[width=0.95\textwidth,height=0.5\textwidth]{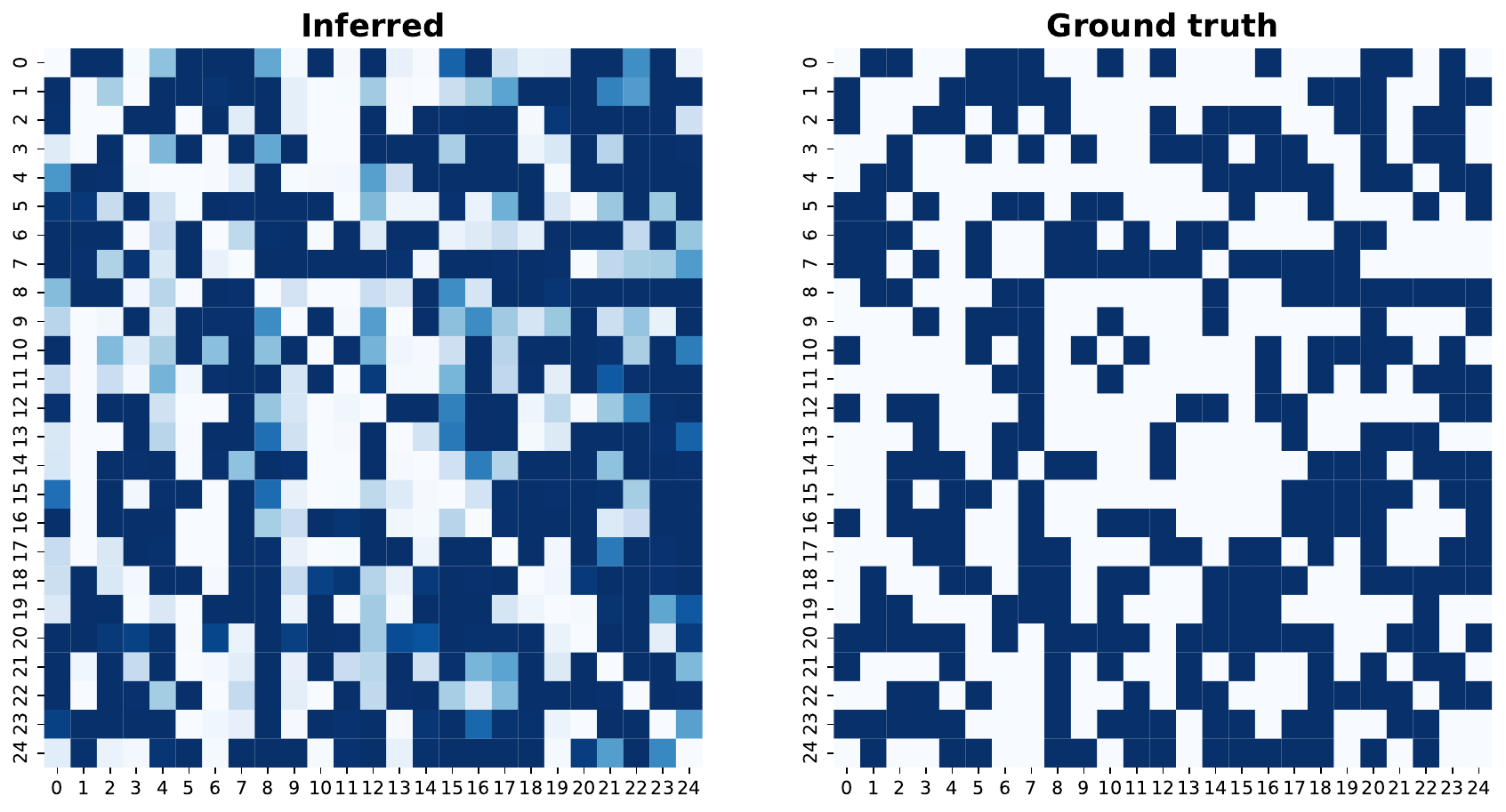}
  \caption{An example of DiffRI inference results and ground truth connection matrix on a 25-node Kuramoto dataset. The left one shows raw inference results. The right one shows the ground truth connection matrix.}
  \label{fig:appen:example}
\end{figure*}

As demonstrated in Fig.~\ref{fig:appen:example}, DiffRI can accurately infer interactions in the 25-node Kuramoto datasets.

\begin{figure*}[htbp]
  \centering
  \includegraphics[width=\textwidth,height=0.5\textwidth]{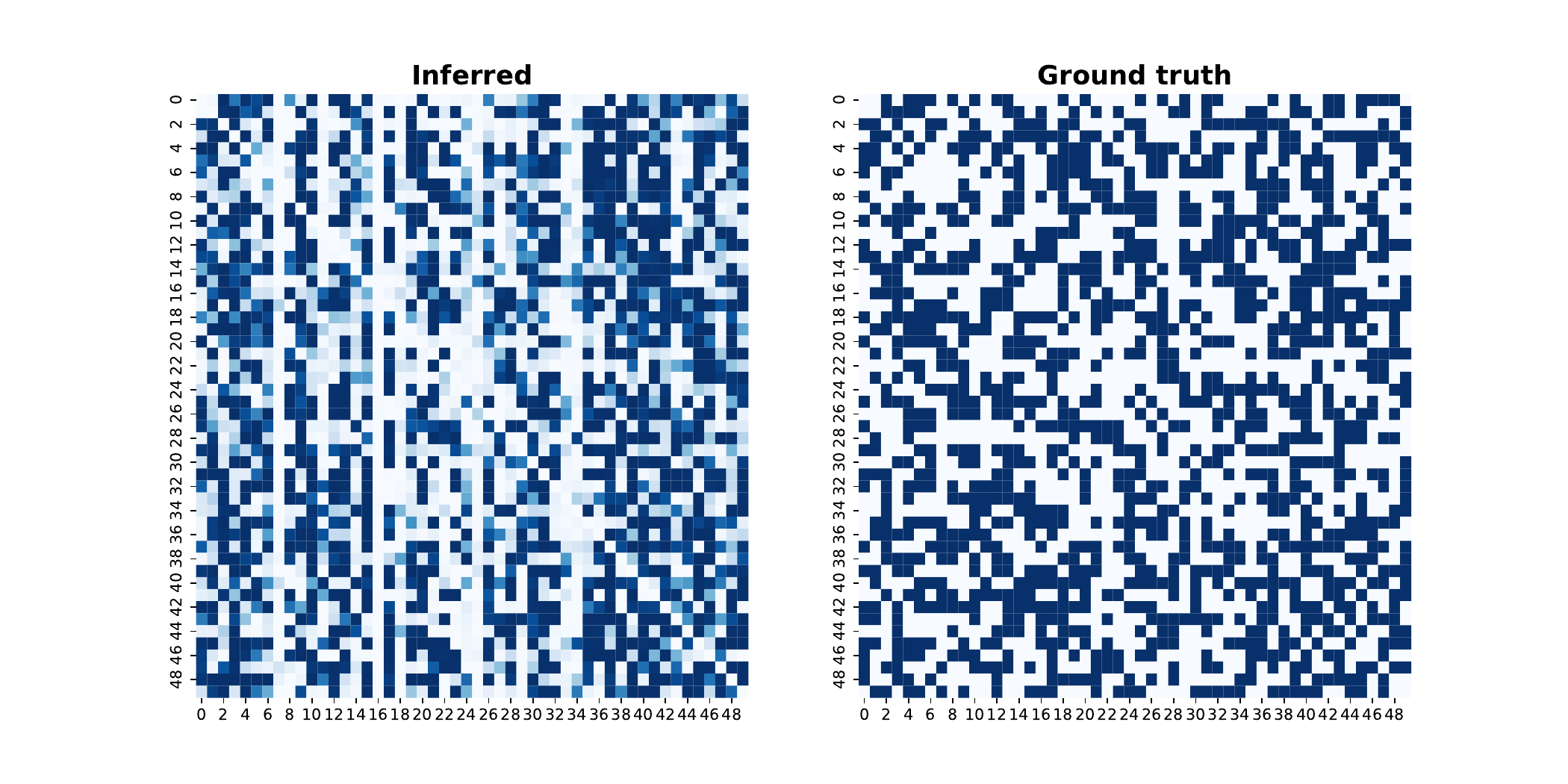}
  \caption{An example of DiffRI inference results and ground truth connection matrix on a 50-node Kuramoto dataset.}
  \label{fig:appen:example:50}
\end{figure*}

We further perform experiments on a 50-node Kuramoto dataset comprising 2000 training samples, 400 validation samples, and 400 testing samples. We increase the batch size to 32. Other hyper-parameters are maintained the same. Results over 4 trials with different random number seeds indicate \textit{an average accuracy of $82.3\%$ with $3.7\%$ std and an average AUROC of 0.834 with 0.025 std}. An example illustration is given in Fig.~\ref{fig:appen:example:50}. 

In addition, we plot GPU running time per epoch across different numbers of nodes. We consider 500 samples for every system, with a batch size of 64. Every sample has a size of $K\times L$, where $K$ equals the number of nodes and $L$ is the time length (=100).

\begin{figure*}[htbp]
  \centering
  \includegraphics[width=0.65\textwidth,height=0.48\textwidth]{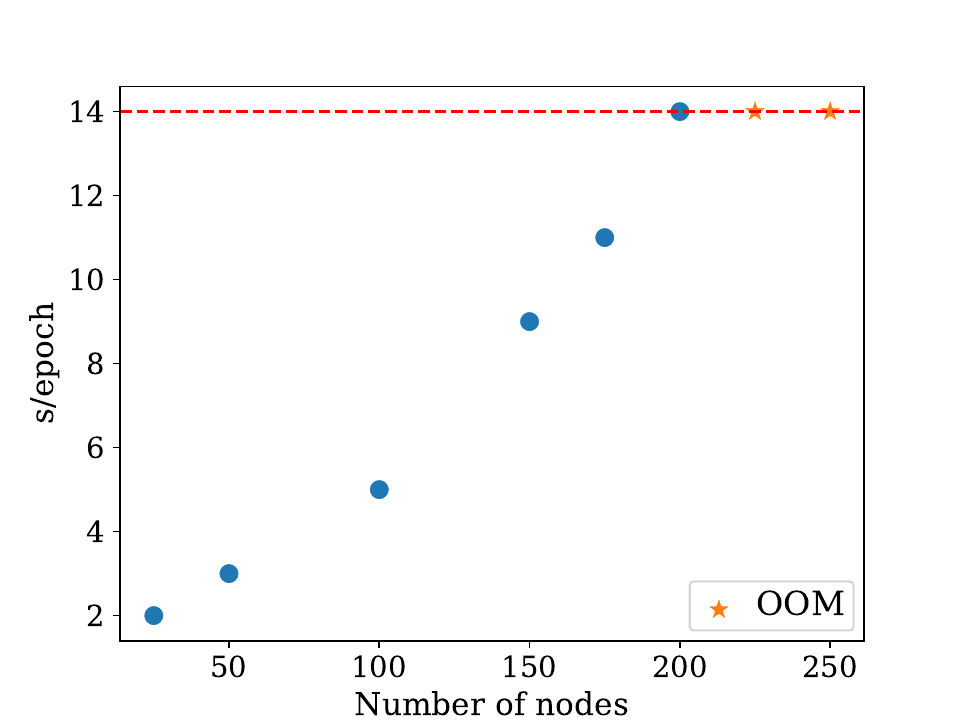}
  \caption{Training time per epoch with scaled systems.}
  \label{fig:appen:computationalScaled}
\end{figure*}

\subsection{Experiments with downstream imputation tasks}
\label{appen:downstream}
We present the performance of DiffRI on the downstream task. Initially, we train DiffRI with varying missing data ratios to simulate real-world situations where some information is never available. We then test the model by imputing 50\% of the remaining data and calculating RMSE and MAE values. As this paper does not primarily focus on downstream tasks, only results on DiffRI are included.

% \usepackage{multirow}
\begin{table}[htbp]
\caption{\label{tab:downstream} DiffRI's performance on imputation tasks. MR=Missing Ratio. Kura=Kuramoto}
\begin{adjustbox}{max width=\textwidth}
\begin{tabular}{cllllll}
\hline
\multirow{2}{*}{\textbf{Metric}} & \multicolumn{2}{c}{\textbf{MR=0\%}} & \multicolumn{2}{c}{\textbf{MR=25\%}} & \multicolumn{2}{c}{\textbf{MR=50\%}} \\
                                 & Spring-10              & Kura-10           & Spring-10              & Kura-10            & Spring-10              & Kura-10            \\ \hline
\multicolumn{1}{l}{RMSE (MAE)}   & 0.110 (0.032)          & 0.127 (0.036)         & 0.154 (0.070)          & 0.167 (0.065)          & 0.164 (0.071)          & 0.256 (0.131)          \\ \hline
\end{tabular}
\end{adjustbox}
\end{table}

\subsection{Ablation studies on the implementation of feature interaction layers}
\label{appen:ablation_on_feature_interaction}
We compare two implementations of the feature interaction layer, using MLP and LSTM. In the Kuramoto 5/10 nodes and Spring 5/10 nodes systems, we show that LSTM constantly performs much better than MLP, thus partially justifying our model architecture design. This is reasonable as LSTM contains inductive bias for time series data.

\begin{table}[htbp]
\caption{\label{tab:ablation_feature_inter} Inference accuracy comparisons between LSTM and MLP implementations}
\begin{adjustbox}{max width=\textwidth}

\begin{tabular}{lllll}
\hline
\textbf{Implementations} & \textbf{Spring-5 nodes} & \textbf{Spring-10 nodes} & \textbf{Kuramoto-5 nodes} & \textbf{Kuramoto-10 nodes} \\ \hline
\textbf{MLP}             & 84.1 (7.9)              & 68.0 (6.4)               & 75.4 (17.5)               & 80.4 (5.2)                 \\
\textbf{LSTM}            & \textbf{97.5 (5.6)}     & \textbf{98.0 (2.0)}      & \textbf{92.6 (8.4)}       & \textbf{95.3 (3.7)}        \\ \hline
\end{tabular}
\end{adjustbox}
\end{table}

\section{Limitations}

While we have shown DiffRI's effectiveness and potential in the relational inference of time series data, there are several limitations that are worth attention.

First, DiffRI is designed for \textit{transductive} relational inference tasks. Therefore, whenever new nodes or new edges appear, the model needs to be retrained. We leave the design of an \textit{inductive} version of DiffRI as an interesting future direction. A rough solution is combining inductive graph learning with the diffusion imputation model.

Second, although we designed our model to infer directed edges and respect causal, we found that DiffRI tends to confuse causal directions in experiments. This point needs to be improved for causal inference. 

Third, we assume the underlying graph structure is static along temporal evolution. This assumption might not hold true in some scenarios. We address dynamic connectivity as a future research direction. Basically, a dynamic connectivity inference requires the inferred edge $\mathbf{E}_t^k$ to become time-dependent. In DiffRI, we could moderate the edge prediction module to give $\mathbf{E}_t^k$ for every $t=1,...,T$. Correspondingly, the feature interaction layer module needs to update $\mathbf{x}_\text{feat}^k$ by utilizing $\mathbf{E}_t^k$ step by step temporally. 

\section{Social impacts}
This paper presents work whose goal is to advance the field of Machine Learning. There are many potential societal consequences of our work, none which we feel must be specifically highlighted here.

\end{document}